\documentclass[10pt,twocolumn,letterpaper]{article}

\usepackage{iccv}
\usepackage{times}
\usepackage{epsfig}
\usepackage{graphicx}
\usepackage{amsmath}
\usepackage{amssymb}

\usepackage{graphicx}
\usepackage{amsmath}
\usepackage{amssymb}
\usepackage{booktabs}
\usepackage{amsmath, amsfonts, times, epsfig, amssymb,  blindtext, float, subcaption, multicol, array, multirow, graphicx, etoolbox,  accents, tabularx, tabu, comment, soul}
\usepackage{color}
\usepackage{booktabs}
\usepackage{colortbl}
\usepackage{arydshln}

\def\etal{\emph{et al}.}
\def\ie{\emph{i.e.,}}
\def\eg{\emph{e.g.,}}
\def\sota{\emph{state-of-the-art}}
\definecolor{tablegray}{gray}{.9}

\usepackage{lipsum}

\usepackage[pagebackref=true,breaklinks=true,letterpaper=true,colorlinks,bookmarks=false]{hyperref}

\iccvfinalcopy 

\def\iccvPaperID{8} 

\ificcvfinal\fi

\begin{document}

\title{Denoising Diffusion for 3D Hand Pose Estimation from Images}

\author{Maksym Ivashechkin, Oscar Mendez, Richard Bowden\\
University of Surrey, United Kingdom\\
{\tt\small \{m.ivashechkin, o.mendez, r.bowden\}@surrey.ac.uk}
}

\maketitle
\ificcvfinal\thispagestyle{empty}\fi

\begin{abstract}


Hand pose estimation from a single image has many applications.
However, approaches to full 3D body pose estimation are typically trained on day-to-day activities or actions. As such, detailed hand-to-hand interactions are poorly represented, especially during motion. We see this in the failure cases of techniques such as OpenPose \cite{openpose} or MediaPipe\cite{mediapipe}.
However, accurate hand pose estimation is crucial for many applications where the global body motion is less important than accurate hand pose estimation. 

This paper addresses the problem of 3D hand pose estimation from monocular images or sequences. 
We present a novel end-to-end framework for 3D hand regression that employs diffusion models that have shown excellent ability to capture the distribution of data for generative purposes.
Moreover, we enforce kinematic constraints to ensure realistic poses are generated by incorporating an explicit forward kinematic layer as part of the network. 
The proposed model provides state-of-the-art performance when lifting a 2D single-hand image to 3D. However, when sequence data is available, we add a Transformer module over a temporal window of consecutive frames to refine the results, overcoming jittering and further increasing accuracy.

The method is quantitatively and qualitatively evaluated showing state-of-the-art robustness, generalization, and accuracy on several different datasets.
\end{abstract}
\section{Introduction}
\vspace{-0.2cm}
\label{sec:intro}
Accurate 3D human pose estimation from a single image is a challenging problem that must overcome  
image quality, occlusions, motion blur, hand interaction, etc.
The problem is often tackled by decomposition into separate body and hand pose reconstruction stages.
Body pose estimation has seen significant advancements, with numerous solutions proposed in both the academic literature and available as open-source implementations.
However, accurate hand estimation remains a challenge.

The commonly used~{\sota} estimators such as MediaPipe, OpenPose, MMPose~\cite{mediapipe,openpose,mmpose2020} were trained on large-scale datasets and have good generalization for detecting human joints in the image, especially for the human body.
Nevertheless, 2D hand estimation is not always accurate, and often completely fails in the presence of a motion blur or hand-to-hand interaction, while 3D estimation from 2D is even less reliable.




The decision to separate hand and body pose estimation is motivated by several factors. Firstly, full-body reconstruction necessitates higher image resolutions due to the larger size of the body in the image. However, the distribution of hand points differs from that of the body, as they are denser and in closer proximity. Additionally, the relatively small size of the hand makes it impractical to estimate both parts simultaneously. Consequently, most methods in the literature approach 3D hand and body pose estimation independently. In this work, we specifically focus on addressing the more challenging and crucial task of 3D hand pose estimation.

\subsection{Related work}
\vspace{-0.15cm}
Hand pose estimation from a single image has been extensively studied in the literature for several years, with various approaches focusing primarily on leveraging deep learning and convolutional techniques to process images.
These approaches aim to tackle the problem through either direct image-to-3D estimation or a two-step approach involving image-to-2D and 2D-to-3D methods.

Moon~{\etal}~\cite{Moon_2020_ECCV_InterHand2.6M} propose InterHand2.6M -- a large hand image dataset with complex hand-to-hand interactions, and a baseline method that by utilizing ResNet\cite{resnet} from image input, predicts a 3D Gaussian heatmap for image coordinate and relative depth regression.
The final 3D coordinates are obtained via back-projecting points using normalized camera intrinsic parameters and absolute depth estimated by the RootNet~\cite{Moon2019CameraDT}.

Spurr~{\etal}~\cite{spurr_cross-modal} employ a statistical approach to correlate input images with 3D pose embeddings. It exploits an RGB image encoder (ResNet) and decoder, along with separate encoders and decoders for the 3D pose. Three encoder-decoder pairs are trained: image-to-image, image-to-3D, and 3D-to-3D. The primary pair consists of the image encoder and 3D pose decoder, while additional pairs contribute to regularizing the latent embedding space.
Yang~{\etal}\cite{disentagling_hands_linlin} propose a method similar to\cite{spurr_cross-modal} that utilizes a latent space for image synthesis.
However, their approach disentangles the embedding space into independent factors and introduces an additional latent variable.

Zimmerman~{\etal}~\cite{zb2017hand} first estimate the 2D keypoints of a hand and then regress the 3D pose from these keypoints in the canonical frame. The hand orientation is separately determined by predicting a single rotation matrix.
Additionally, the authors have provided a rendered hand dataset (RHD) consisting of synthetic hand poses.
PeCLR, proposed by Spurr~{\etal}~\cite{spurr2021self}, employs a contrastive loss on image pairs with diverse augmentations. The network maximizes agreement between identical images with varied augmentation while minimizing agreement with dissimilar images. Using image features extracted by ResNet, the network predicts 2.5D keypoints ({\ie} image coordinates and relative depth), and the 3D pose is obtained by back-projecting.

The hand estimation literature encompasses various methods that utilize the MANO library~\cite{MANO:SIGGRAPHASIA:2017} for hand parameterization, particularly focusing on volumetric hand prediction.
Guan~{\etal} introduce MobileHand~\cite{MobileHand:2020}, a model that predicts camera rotation, camera scale, camera translation, joint hand angles, and shape to generate a MANO hand mesh. 
Similarly, Boukhayma~{\etal}~\cite{boukhayma20193d} present an end-to-end method for combined 3D hand with mesh estimation from images and 2D heatmaps.
Kulon~{\etal}~\cite{Kulon_2020_CVPR} propose a weakly supervised approach for 3D hand pose estimation. The authors extract 2D keypoints by running OpenPose on images and optimize the MANO hand model to align the projection of 3D points with the OpenPose 2D keypoints. The method exploits a ResNet image encoder to process the input image, followed by a convolutional decoder that predicts the hand mesh by sampling the neighborhood constructed with the spiral operator.




Interactions between hands raise a significant challenge and have been extensively explored in the body of research.
Wang~{\etal} introduce RGB2Hands~\cite{wang_rgb2hands}, a comprehensive framework that addresses the estimation and tracking of interacting 3D hands from video inputs.
The method leverages various information sources, including hand segmentation, depth data, image points, vertex-to-pixel mapping, and hand-to-hand distance, fusing them to regress MANO hand parameters.
The hand interaction problem was also approached by Fan~{\etal}~\cite{fan2021digit} who propose a method for 3D interacting hands prediction from a monocular image by extracting visual and semantic features via CNN. Furthermore, by utilizing a segmentation probability mask, the method regresses 2.5D coordinates and recovers the 3D pose through inverse perspective projection.
Recent works exploring hand interactions and incorporating the MANO model are also presented in~\cite{Zhang2021InteractingT3,Li2022intaghand,meng2022hdr}.

Diffusion models have recently proved themselves as an efficient method for model training, and they are able to generate high-quality samples,~{\eg} images.
They have outperformed~{\sota} generative models such as generative adversarial networks, variational autoencoders, etc.
The denoising diffusion model as a parameterized Markov chain was presented by Ho {\etal}~\cite{ho_denoising_diffusion}.
Additionally, the diffusion models with more improvements were studied in variational diffusion models~\cite{NEURIPS2021_b578f2a5} by Kingma~{\etal}, simple diffusion~\cite{simple_diffusion} of Hoogeboom~{\etal}, improved denoising diffusion probabilistic models~\cite{improved_denoising} by Nichol~{\etal}, etc. 

Several methods have employed diffusion models for 3D body pose estimation by utilizing 2D input keypoints.
Holmquist~{\etal} introduce DiffPose~\cite{diffpose}, which uplifts a human body from 2D to 3D using a conditional diffusion model. The approach involves extracting heatmaps of body joints from an input image and converting them into joint-wise embeddings used for conditioning. While DiffPose demonstrates promising results in body pose evaluation, the authors acknowledge the limitation of its two-step approach, which disregards some information from the image features.
Another method that incorporates diffusion models is DiffuPose by Choi~{\etal}~\cite{diffupose}. This approach performs 2D to 3D body pose uplift by conditioning the diffusion model with 2D keypoints obtained from an off-the-shelf 2D detector. Additionally, DiffuPose replaces the commonly used U-Net module for noise prediction with a graph convolutional network.
Gong~{\etal}~\cite{gong2023diffpose} leverage the diffusion model to recover a true distribution of 3D body poses by conditioning it with spatial context information from 2D points.


Zheng~{\etal} introduce PoseFormer~\cite{zheng20213d}, where authors explore the application of temporal transformer models for pose estimation. The approach incorporates both spatial and temporal information in the transformer to generate a 3D pose estimation for a middle frame using a sequence of 2D estimates.
Furthermore, Jiang~{\etal} present Skeletor~\cite{JIANGTAO2021SSTf}, a sequence-to-sequence model that leverages the encoder part of the transformer to refine 3D poses obtained from 2D.



\begin{figure*}
    \centering
    \includegraphics[width=0.94\linewidth]{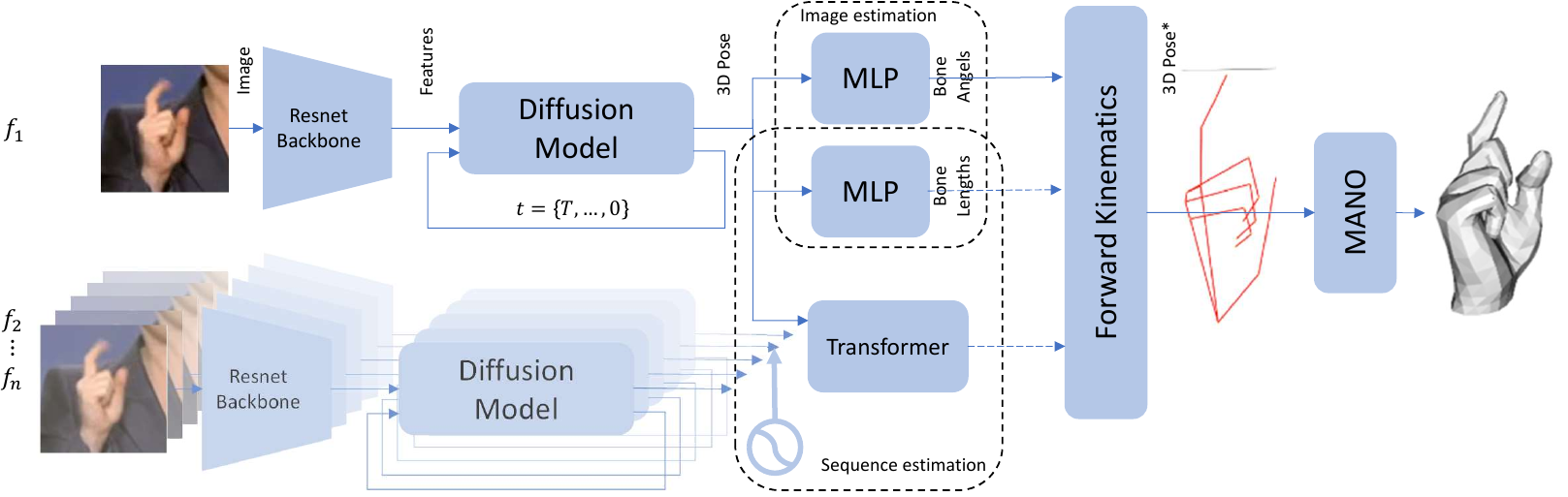}
    \vspace{-0.3cm}
    \caption{This pipeline shows our 3D hand pose estimation denoising diffusion model. Input images are processed by a CNN ({\eg} ResNet), the features then condition the diffusion model. The U-Net~\cite{unet} model starts from pure Gaussian noise and denoises the 3D points iteratively. The inverse kinematics MLP module takes a noisy 3D pose and predicts angles and bone lengths that are fused in the forward kinematics to return a realistic pose. The transformer refines the diffusion model estimate over a sequence. Finally, hand angles and bone lengths are used to produce a hand mesh model~\cite{MANO:SIGGRAPHASIA:2017}. 
    }
    \label{fig:diffusion_model}
\end{figure*}
Within the literature, the most common approach is to extract 2D keypoints, and then uplift them to 3D space.
However, the detection of the 2D joints of the hands is in itself challenging for several reasons.
Firstly, the hands are much smaller than the body which makes them more difficult to detect.
Secondly, hands can move significantly faster than other body parts, hence motion blur often occurs in the image.
Finally, hand interactions often introduce self-occlusion, further complicating the detection process.

The widely adopted approach has two steps, wherein the first step runs a convolutional neural network (CNN) to detect 2D human keypoints on the input image, and then, for instance, a multi-layer perceptron (MLP) takes the 2D joints and outputs the 3D pose.
The benefit of such methods is a two-step decomposition and ease of generalization in the uplift stage.
However, the main issue of this method is its complete reliance on the accuracy of the image detector.
Training a CNN detector on one dataset may provide correct keypoint predictions for images in that dataset, but in practice, the detector can struggle to generalize to images outside the training data,~{\eg} with a different background, image noise, clothes, etc.
\emph{State-of-the-art} 2D hand detectors such as MediaPipe or OpenPose often completely fail or output inaccurate 2D joints in the presence of motion blur and hand-to-hand interaction. If 2D detection fails, then uplifting to 3D is impossible.

\subsection{Motivation}
\vspace{-0.2cm}
To overcome the limitations of the traditional two-step approach and mitigate error propagation, methods that predict 3D pose directly from image features could be utilized. However, training such direct approaches can be more challenging due to the complex mapping between 2D images and 3D poses. This is where diffusion models offer a compelling solution.
Diffusion models have recently emerged as a promising approach for pose estimation tasks conditioned on 2D information. These models excel at denoising and capturing complex data relationships, making them well-suited for the task of regressing a 3D pose from image features. By leveraging the denoising capability of diffusion models, they can effectively handle noise and uncertainty in the input data, resulting in more robust and accurate pose estimations.
Incorporating an inverse kinematics layer further enhances the effectiveness of diffusion models for hand pose estimation. By enforcing kinematic constraints, such as joint angles and joint limits, the model can generate more realistic and anatomically plausible hand poses. This not only improves the accuracy of the estimated 3D poses but also ensures that the generated poses adhere to the natural range of motion for human hands.
Compared to the simpler MLP uplift method, diffusion models offer distinct advantages.

Hand pose estimation that suffers from the effect of fast motion can actually be to our benefit by integrating temporal information and cues over multiple frames.
Therefore, we propose a temporal model based on a transformer that can leverage this additional temporal information and increase performance further.

We propose a novel hand estimation method that leverages a diffusion model conditioned on image features to directly predict a 3D pose avoiding an explicit two-step approach.
The output pose undergoes an IK (inverse kinematics) layer to enforce physical constraints on a hand, and on a sequence input the additional transformer module eliminates jittering to ensure smoothness of estimate that could be converted to mesh representation afterwards.  



\vspace{-0.2cm}

\section{Methodology}
\label{sec:methodoly}

\begin{figure}[t]
    \centering
    \includegraphics[width=0.40\linewidth]{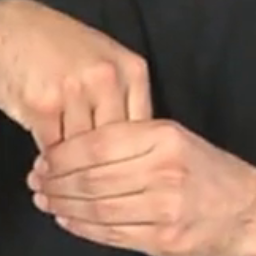}
    \includegraphics[width=0.44\linewidth]{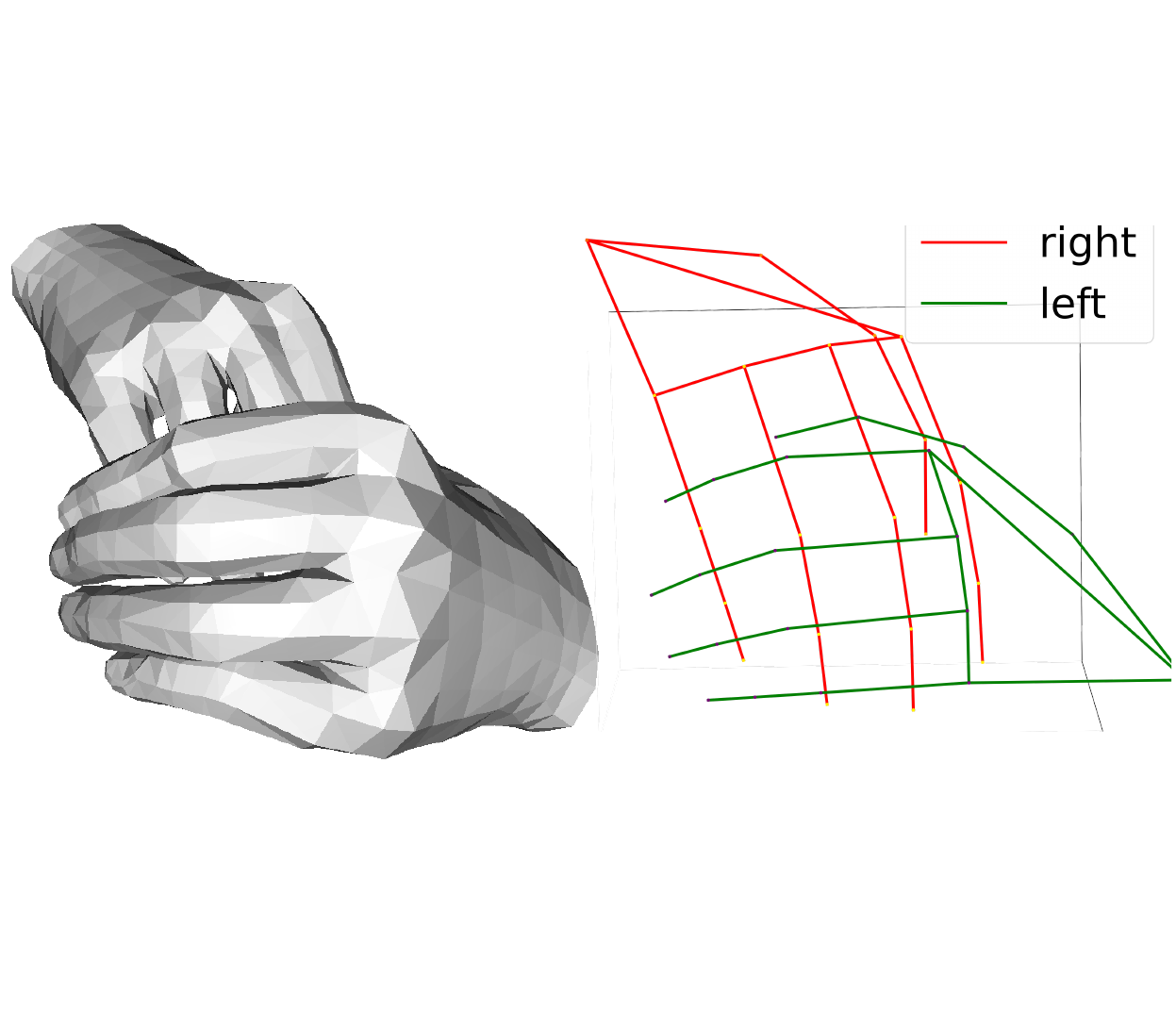}
    \includegraphics[width=0.44\linewidth]{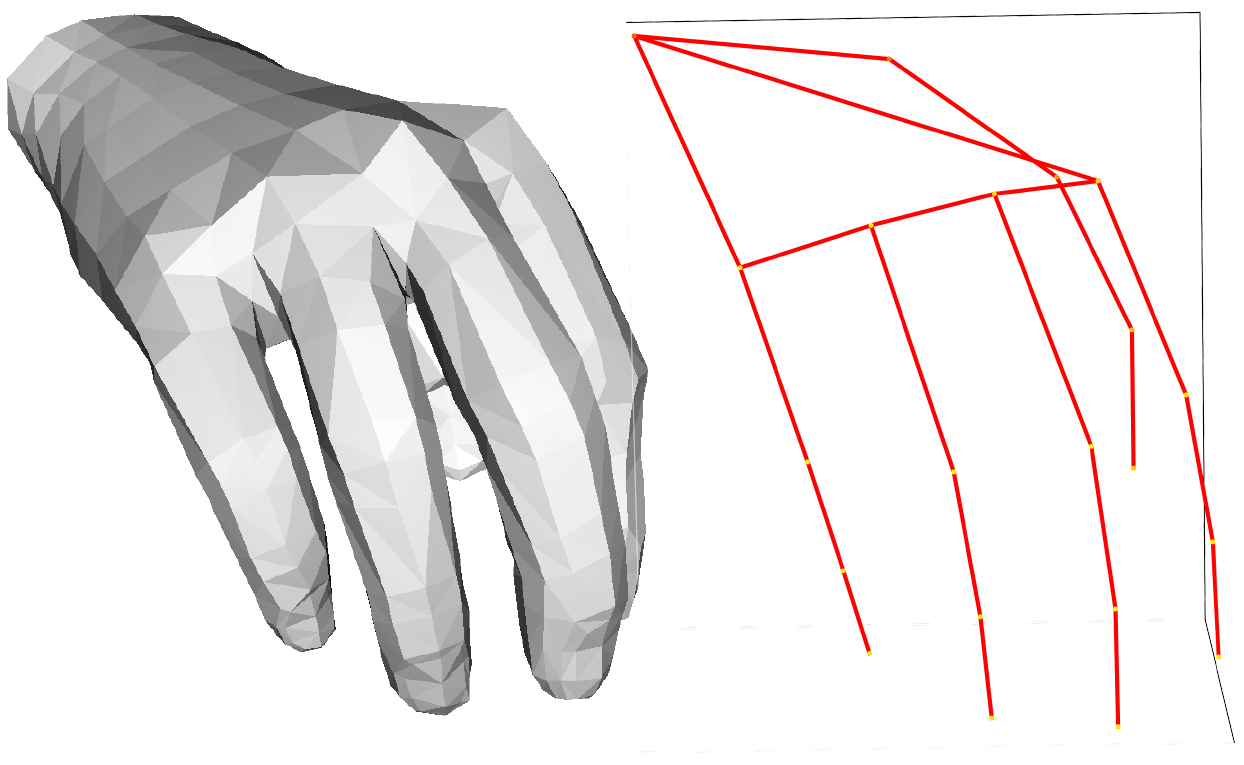}
    \includegraphics[width=0.44\linewidth]{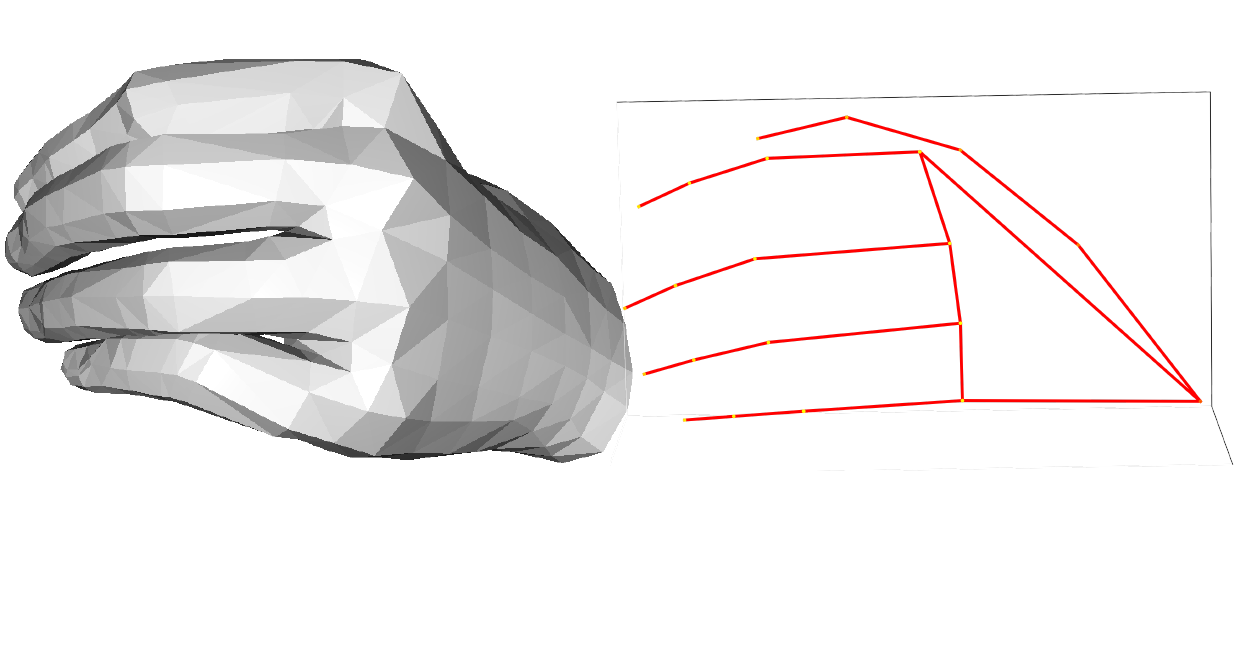}
    \vspace{-0.3cm}
    \caption{The proposed model is able to predict right and left hands from a single image of hands interaction. In the top right figure, the estimated 3D hand interaction is shown together with a hand mesh. The bottom figures show individual right and left hands' skeletons with mesh. The random hand interaction image was not part of the training set.}
    \label{fig:hands_interaction}
\end{figure}
\vspace{-0.3cm}
The overview of our pipeline is provided in Figure~\ref{fig:diffusion_model}. The model consists of a pre-trained ResNet~\cite{resnet} and a diffusion model that predicts the 3D pose of the hand based on the ResNet features. For a static image, two MLP modules then predict the parameters that are fed into the Forward Kinematics (FK) layer, namely the bone lengths and the bone angles. For a temporal sequence, we feed the output of the diffusion model into a small Transformer which predicts the bone angles using the temporal context. However, as the bone lengths remain consistent over the sequence of consecutive frames, these are still estimated by the MLP layer. We now describe each step in turn.

\subsection{Feature Extraction}
\vspace{-0.2cm}
ResNet features  are taken after average pooling, additionally, we stack one hidden fully-connected layer to decrease the feature dimensionality.
Without loss of generality, the ResNet feature network is applied only to ``right'' hand images.
More explicitly, we feed the network right-hand images and horizontally flipped left-hand images.
The output of left-hand images is then flipped back by multiplying the $x$-axis of the output 3D joints with -1.
This formulation is also applicable to hands' interactions, see Figure~\ref{fig:hands_interaction}).

\subsection{Bone Prediction}
\vspace{-0.2cm}
Bone angle prediction is decomposed into two parts.
The first one predicts the rotation in the camera frame,~{\ie} root joint (wrist)  orientation, and the second estimates the angles of other joints.
The reason for separation is that root rotation is unconstrained and therefore a high-dimensional parameterization can be used,~{\eg} 9 DoF singular value decomposition orthogonalization~\cite{learning_3d_skeleton}. 

To enforce the angular and bone length constraints, the bone 
angles and length prediction
are followed by a sine normalization function to clamp the values from -1 to 1, transform to a 0-1 range, and finally multiplied by constraints as follows:
\begin{equation}
    a = \frac{\sin(x) + 1}{2} (a_{\max} - a_{\min}) + a_{\min},
\end{equation}
where $x$ is a neuron output, and $a \in [a_{\min}, a_{\max}]$ is the constrained angle.

\subsection{Hand Mesh}
\vspace{-0.1cm}
The hand mesh is generated independently using the MANO model~\cite{MANO:SIGGRAPHASIA:2017}.
Note: we use MANO only for visualization, it is not integrated into the network for training.
MANO uses angles, shape parameters, and pre-trained weights to generate a mesh combining the forward kinematics. 
The angles returned by the proposed model are used to initialize the MANO convention.
We prioritize our own hand model as it enables us to parameterize the hand with specific degrees of freedom, Euler angles, and constraints for joint articulation and finger lengths, while MANO relies on statistically precomputed hand shapes.


\subsection{Model Supervision}
\vspace{-0.2cm}
The model is supervised via the ground truth 3D poses in the camera frame.
We assume a perspective camera model with a projection matrix $\mathbf{P} = \mathbf{K} \:[\mathbf{R} \:|\: \mathbf{t}]$.
Where $\mathbf{K} \in \mathbb{R}^{3\times3}$ is an intrinsic matrix, and $(\mathbf{R}, \mathbf{t})$ are the camera's rotation and translation that transform an object from the world to camera frame.
The 3D hand with $N$ joints in the world frame is a matrix $\mathbf{X}_W$ of size $3\times N$ where points are stacked in columns.
The pose in the camera frame used for model supervision is hence, $\mathbf{X}_C = \mathbf{R} \mathbf{X}_W + \mathbf{t}\,\mathbf{1}_N^\top$, and its 2D projection onto the image plane is $\mathbf{X}_I \sim \mathbf{K} \mathbf{X}_C$.

The network does not predict the origin of the pose in 3D space and the bone length scale, because this is challenging for a single image input.
Therefore, the training is invariant to hand scale and origin.

We incorporate several loss functions with suitable weighting that have empirically proved to generate more accurate models.
The first loss is the mean absolute error ($L_1$ loss) between the ground truth pose and the estimated one,~{\ie} $|\mathbf{X}^*_C$ - $\hat{\mathbf{X}}_C|$.
The second is additional supervision with corresponding image points (if a dataset contains intrinsic),~{\ie} $||\mathbf{X}^*_I - \hat{\mathbf{X}}_I||$.
An MLP that estimates a rigid hand rotation enables prediction of the 3D hand pose in both the canonical and camera frames, hence, this decomposition suits being trained separately applying a 3D loss on both canonical and camera frame 3D output.
Finally, the contrastive loss (SimCLR~\cite{contrastive_framework}) is applied to image features from a ResNet extracted from differently augmented image pairs to maximize agreement on the same hand images and minimize on different pairs as suggested in PeCLR~\cite{spurr2021self}.
Images of the hands are augmented with different levels and types of image noise, blur, sharpening, jittering, etc., and the hand location in the images is also randomly shifted and scaled.


\subsection{3D Diffusion}
\begin{figure}
    \centering
    \includegraphics[width=0.99\linewidth]{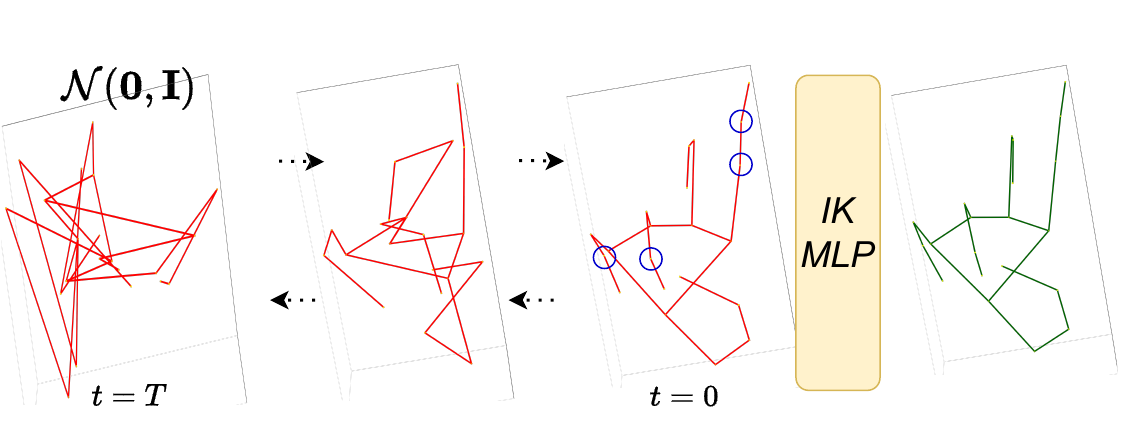}
    \vspace{-0.35cm}
    \caption{Denoising steps for 3D pose estimation as part of the network pipeline demonstrated in Figure~\ref{fig:diffusion_model}. The output 3D hand from the diffusion module (in red) is perturbed by noise and does not look realistic (see blue circles on the hand pose highlighting the inaccuracies), while the green hand returned by the IK MLP module has preserved hand constraints.}
    \label{fig:denoising_steps}
\end{figure}
\vspace{-0.15cm}
Diffusion models are a class of generative models that are used to predict high-quality samples by gradual denoising.
It is common to distinguish \textit{forward} and \textit{reverse} diffusion processes.
Let us denote a data vector ${\bf x}_0 \sim q({\bf x})$ sampled from a real distribution $q$, and $T$ as the number of time steps where Gaussian noise with variance $\bigl\{\beta_t \in (0, 1)\bigr\}^T_{t=1}$ is added to vector ${\bf x}_0$ at each step $t\in[0, T]$  to generate a sequence of samples $\bigl\{{\bf x}_t\bigr\}^T_{t=1}$.
The latent variable ${\bf x}_t$ is then sampled from distrubtion $q({\bf x}_t | {\bf x}_{t-1})$ of mean $\boldsymbol\mu_t $ and variance $\mathbf{\Sigma}_t$ as follows:
\begin{equation}
{\bf x}_t \sim q({\bf x}_t | {\bf x}_{t-1})=\mathcal{N}({\bf x}_t; \sqrt{1 - \beta_t}\, {\bf x}_{t-1}, \beta_t\mathbf{I}).
\end{equation}
Given a useful property of reparameterization~\cite{kingma2014autoencoding} to obtain ${\bf x}_t$, it is enough to sample it from the following distribution:
\begin{equation}
{\bf x}_t \sim q({\bf x}_t | {\bf x}_0) = \mathcal{N}({\bf x}_t; \sqrt{\bar\alpha_t}\, {\bf x}_0, (1 - \bar\alpha_t)\mathbf{I}),
\end{equation}
where $\bar\alpha = \prod^t_{i=1}(1 - \beta_i)$.
This enables a \textit{forward} process to efficiently noise the input data ${\bf x}_0$ to a certain time-step $t$.

In the \textit{reverse} process, the goal is to obtain ${\bf x}_0$ from ${\bf x}_T$, which for $T\to\infty$ steps is close to an isotropic Gaussian distribution.
The reverse distrubution $q({\bf x}_0 | {\bf x}_t)$ is unknown, while the distrubution $q({\bf x}_{t-1} | {\bf x}_t)$ is very difficult to obtain.
Therefore, the diffusion model approximates $q({\bf x}_{t-1} | {\bf x}_t)$ via a parameterized neural network to learn a conditional probability $p_{\theta} ({\bf x}_{t-1} | {\bf x}_t)$ to estimate this joint probability $p_\theta ({\bf x}_{0:T})$ ({\ie} reverse process), which is defined as a Markov chain with learned Gaussian transition~\cite{ho_denoising_diffusion}
\begin{equation}
p_\theta({\bf x}_{0:T}) = p({\bf x}_T) \prod^T_{t=1}p_\theta({\bf x}_{t-1} | {\bf x}_t),
\end{equation}
where $p({\bf x}_T) \sim \mathcal{N}(\mathbf{0}, \mathbf{I})$.
Essentially, the neural network is trying to predict the mean $\boldsymbol\mu_\theta({\bf x}_t, t)$ and variance $\boldsymbol\Sigma_\theta({\bf x}_t, t)$ of the distribution $p_\theta({\bf x}_{t-1} | {\bf x}_t)$ conditioned on time-step $t$.

For the hand pose estimation problem, the data sample corresponds to a set of 3D hand points,~{\ie} ${\bf x}_0 \in \mathbb{R}^{3N}$, where $N$ is a number of hand points ({\eg} 21 in the experiments).
Additionally, we condition the denoising model $p_\theta$ not only on the time-step $t$ but also on the extracted image features ${\bf f}$ to provide the network information about the corresponding images.
The most common architecture for denoising diffusion models is a U-Net, modified to have the time-embeddings of each time-step $t$.
In the experiments, we use a 1D U-Net which takes 3D points concatenated with image features and time-step information.

For training diffusion models, the objective is normally to minimize the Kullback-Leibler divergence~\cite{Joyce2011}.
Hovewer, Ho~{\etal}~\cite{ho_denoising_diffusion} made simplifications, first by setting ${\bf \Sigma}_\theta({\bf x}_t, t) = \sigma^2_t\mathbf{I}$, where $\sigma^2 \approx \beta_t$.
Additionally, by exploiting the distribution $q({\bf x}_{t-1} | {\bf x}_t, {\bf x}_0) = \mathcal{N}({\bf x}_{t-1}; \Tilde{\boldsymbol\mu}_t({\bf x}_t, {\bf x}_0), \Tilde{\beta}_t\mathbf{I})$, which is tractable when conditioned on ${\bf x}_0$, Ho~{\etal} suggest to train the reverse process mean function approximator $\boldsymbol\mu_\theta$ to predict $\Tilde{\boldsymbol\mu}_t$.
Consequently, the training of the denoising model is done by taking a gradient descent step on the difference between sampled and predicted noise as follows:
\begin{equation}
    \nabla_\theta ||\boldsymbol\epsilon - \boldsymbol\epsilon_\theta(\sqrt{\bar\alpha_t}{\bf x}_0 + \sqrt{1 - \bar\alpha_t}\boldsymbol\epsilon, t, {\bf f})||^2,
\label{eq:loss_fnc_diff}
\end{equation}
where, $\epsilon \sim \mathcal{N}({\bf 0}, {\bf I})$ is randomly sampled noise, $t$ is sampled uniformly in random within $0$ to $T$ range, $\boldsymbol\epsilon_\theta$ is, for instance, U-Net model conditioned on time-step $t$ and image features ${\bf f}$ to predict the Gaussian noise.
Luo~{\etal}~\cite{understanding_diff} proved that optimizing (\ref{eq:loss_fnc_diff}) gives better performance for a diffusion model than the original ELBO~\cite{Joyce2011}.

During inference, pure Gaussian noise is concatenated with image features, and given the number of denoising time steps, the U-Net model gradually removes noise between two consecutive time steps.
The recursive equation to produce a 3D skeleton from an image is therefore:
\begin{equation}
    {\bf x}_{t-1} = \frac{1}{\sqrt{\alpha_t}} \Bigl({\bf x}_t - \frac{\beta_t}{\sqrt{1 - \alpha_t}}\boldsymbol\epsilon_\theta({\bf x}_t, t, {\bf f}) \Bigr) + \sigma_t {\bf z},
    \label{eq:diff_inference}
\end{equation}
where ${\bf z} \sim \mathcal{N}({\bf 0}, \mathbf{I})$ if $t > 0$, otherwise ${\bf z = 0}$, and the calcuation starts from ${\bf x}_T\sim\mathcal{N}({\bf 0}, \mathbf{I})$.

The proposed 3D hand estimation pipeline incorporating a diffusion model is outlined in Figure~\ref{fig:diffusion_model}. 
Firstly, an input image is processed by a ResNet to extract features that condition the diffusion model.
Subsequently, using equation (\ref{eq:diff_inference}) and starting from the pure Gaussian noise, the diffusion model returns a 3D hand pose.
To enforce kinematic constraints, we add an inverse kinematics MLP layer that takes the output from the diffusion model and generates angles and bone lengths that are fused into a valid 3D pose.
Figure~\ref{fig:denoising_steps} shows the denoising steps of the diffusion model with the IK refining part.

Diffusion models demonstrate stable training, simple supervision, and good accuracy.
In our experiments, the diffusion model outperforms a baseline MLP, which regresses angles and bone lengths from image features.
However, the detrimental aspect of the diffusion model is significantly longer inference time, where at each time step the denoising model has to be executed,~{\eg} $T=50$ in the experiments.

\subsection{Forward Kinematics}
\begin{figure}[htp]
    \centering
    \includegraphics[width=0.64\linewidth]{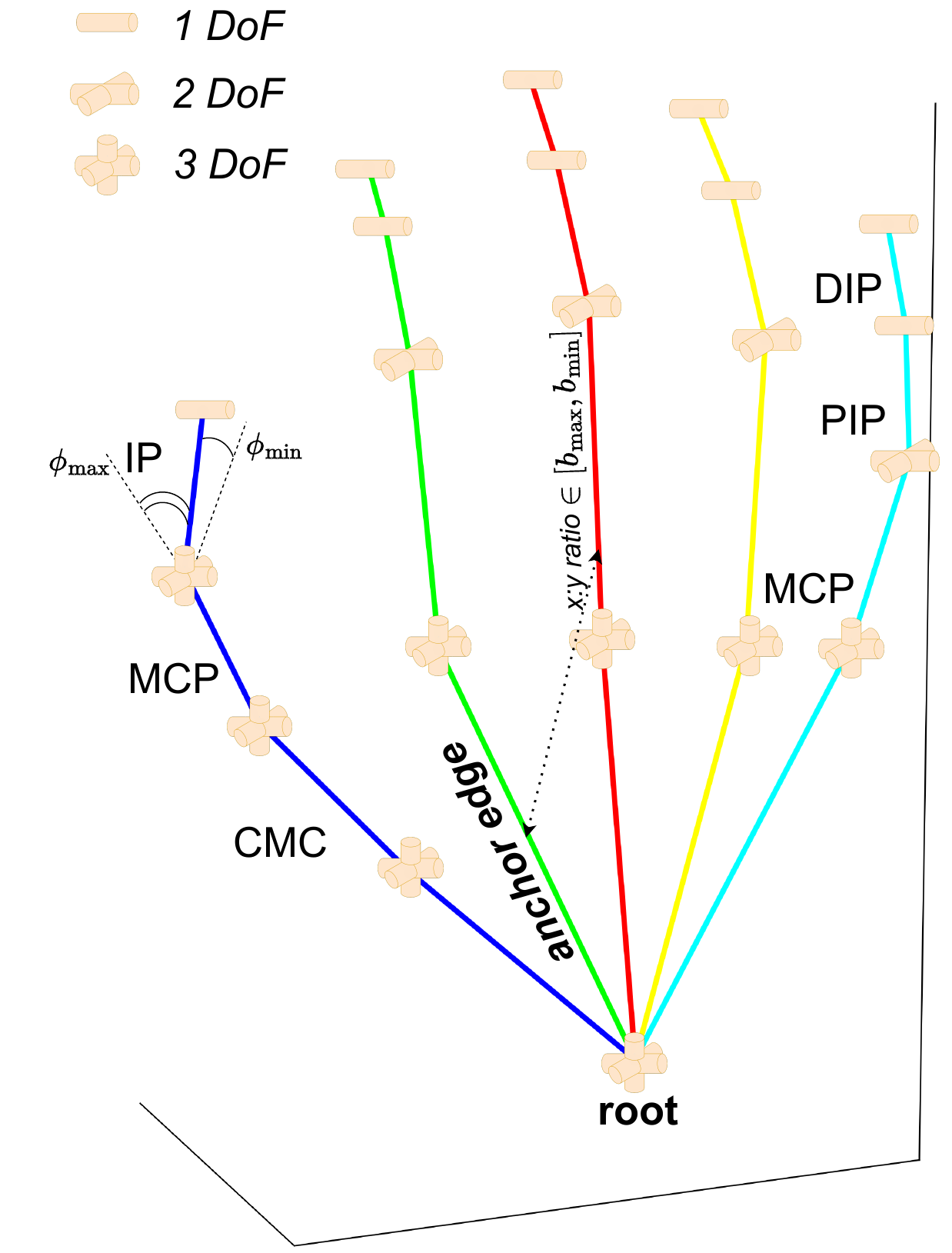}
    \vspace{-0.3cm}
    \caption{The skeleton shows the hand parameterization where each node has assigned degrees of freedom (DoF). The root of the skeleton is  the wrist of the hand. The main edge of the skeleton graph is from the root to index MCP (metacarpophalangeal), in which bone length represents the scale of the hand. The other edges are computed as proportion and with respect to the anchor edge, the proportion ratio is constrained to the feasible range of bone length. Moreover, each angle is constrained to be within a limited range.
    }
    \label{fig:skeleton}
\end{figure}
\vspace{-0.1cm}
Accurate and realistic 3D hand pose estimation requires kinematic constraints such as the limitation of joint rotations, bone length symmetry, etc.
Therefore, we parameterize the hand skeleton as a tree graph with a root node ({\ie} wrist joint), the sets of vertices and edges, where vertices represent the hand joints and the edges match the skeleton bones, see Figure~\ref{fig:skeleton}.

Each node of the graph models the constraints of the joint,~{\ie} its orientation, angular constraints, and degrees of freedom (DoF).
For instance, a node that corresponds to the distal interphalangeal joint has only one degree of freedom, since it can only move in one direction, and its angular limit is from zero to approximately 100 degrees.
The rotation of a joint is represented via Euler angles.
Even though Euler angles are not continuous ({\ie} 0 and $2\pi$), they enable us to easily parameterize the rotation of a joint ({\eg} degrees of freedom), and enforce angular limits, which is more challenging with higher dimensional representations ({\eg} quaternions).

The graph edges also represent the individual offsets of the child node to its parent, and the root offset is the pose's origin in the camera frame.
Instead of working with real distances of edges, we select an anchor edge ({\eg} the longest edge in the skeleton), and all other edges are scaled with respect to this edge.
This approach makes the skeleton easily scalable and more intuitive because human hand proportions are relatively constant.
To control the proportion limits, each edge has an assigned tuple of the maximum and minimum ratio or scale for the anchor edge.

The proposed tree graph is directed, and it has a hierarchical structure beginning from the root vertex.
This aims to build a chain of computations for the forward kinematics layer (FK) by traversing from the root to the leaves of the tree.
The orientation of the nodes is therefore relative to its parent, and the rotation of the root is the orientation of the pose in the camera frame.
This graph representation helps to enforce constraints including symmetry or scaling of the hand structure during the FK processing.

\subsubsection{FK Layer}
\vspace{-0.2cm}
The forward kinematics layer is a non-parametric layer of the network that is implemented by traversing a tree graph via the breadth-first search algorithm (BFS)~\cite{Bundy1984}.
This allows us to process all nodes in parallel at each depth level of the tree.
The computation starts from the root node and expands to its children, and it recursively repeats thereafter.

Each node $i$ has Euler angles relative to the parent node  $\mathbf{e}_i \in \mathbb{R}^3$ that is within a limit $[\mathbf{e}^i_{\min}, \mathbf{e}^i_{\max}]$ and translation offset $\mathbf{o}_i \in \mathbb{R}^3$.
The relative rotation of the node is given by converting Euler angles to a rotation matrix $\mathbf{R}^\prime_i = \phi(\mathbf{e}_i)$ via mapping $\phi : \mathbb{R}^3 \rightarrow \mathbb{R}^{3\times3}$.
The relative rotation matrix $\mathbf{R}^\prime$ can have from zero to three DoF depending on the parametrization of the nodes.

The node's rotation in a camera frame is denoted as $\mathbf{R}_i$ and its position in the 3D space is $\mathbf{p}_i$.
The root hence has relative rotation and offset with respect to the camera frame,~{\ie} $\mathbf{R}_0 = \mathbf{R}^\prime_0$ and $\mathbf{p}_0 = \mathbf{o}_0$.
The position and orientation in space for all other nodes are found by a recursive rule at each stage of the BFS tree traverse as follows:
\begin{equation}
    \mathbf{R}_i = \mathbf{R}_j \mathbf{R}^\prime_i \:\:\:\:\:\:\:\:\:\:\:\:\:\:\:\:\:\:\:\: \mathbf{p}_i = \mathbf{R}_i\, \mathbf{o}_i + \mathbf{p}_{j}.
\end{equation}

The edge $(i,j)$ goes from the parent node $j$ to its child $i$.
The position of each joint is dependent on the parent, while the root node remains unconstrained,~{\ie} it has three DoF without limits.

\subsection{Temporal Transformer}
\vspace{-0.2cm}
When a video sequence is available, we replace the angular MLP with an angular Transformer model to exploit temporal information over the sequence of consecutive frames.
This allows us to overcome noisy predictions caused by motion blur or occlusions, and refine the final estimate.
The transformer outputs a sequence of angles that with fixed bone lengths are fused into the forward kinematics layer to generate 3D poses. 

In this work, we explored different variations of the transformer.
Primarily, the decoder part is unnecessary because a target sequence is not always available, ~{\eg} dataset does not contain the ground truth joint angles.
The input for the Transformer can be either a sequence of 3D points or angles since the MLP and diffusion model work with angles and 3D joints respectively.
The diffusion model at the first stage outputs 3D points, therefore, it is better to avoid the computation of angles with the IK module and pass 3D points directly to the transformer to generate a sequence of angles.


The Transformer encoder consists of several layers and heads.
The inputs are embedded into high-dimensional vectors using a fully-connected layer with a sinusoidal positional encoding.
Additionally, we use an encoder mask  where for each batch we randomly hide at most 50\% of the sequence values.
Together with dropout, it forces the encoder to better generalize and learn  temporal information.

For the bone length smoothing an additional temporal model is not required.
We assume that bone lengths do not change over the sequence, so simple averaging  of the MLP predictions is sufficient.

\section{Experiments}
\subsection{Skeletal parameterization}
\vspace{-0.10cm}
For a hand pose skeleton with 21 joints, we used the 26 degree of freedom parameterization from \cite{Samadani2012MulticonstrainedIK}, where three angles are used for the orientation of the wrist (root joint), eight angles for the four metacarpophalangeal (MCP) points that have two DoF, three angles for the thumb MCP, eight angles for the four interphalangeal (PIP) and the four distal interphalangeal (DIP) joints, one angle
for the thumb interphalangeal (IP), and three angles for the thumb carpometacarpal (CMC) points.
Additionally, apart from the angular parameterization of the joints, we include 15 more angles (three per finger) to position fingers on the correct offset from each other.
Therefore, in total one hand has 41 angles and 20 bone length proportions.

We have also implemented an inverse kinematics (IK) solver to fit 3D hand poses by optimizing angles and hand shape.
The solver helps to extract, and statistically compute, the angular constraints and bone length proportions from a dataset.
The obtained limits are used for model training to predict the pose parameters within the desired range.

\subsection{Quantitive evaluation}
\begin{figure*}
    \centering
    \includegraphics[width=0.99\linewidth]{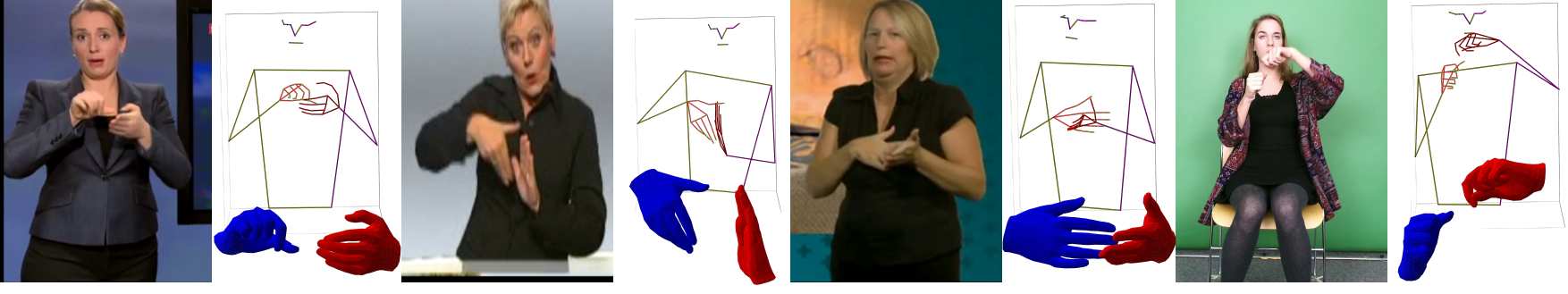}
    \vspace{-0.20cm}
    \caption{Qualitative evaluation of the proposed baseline hand model on four sign language datasets: DSGS~\cite{li-etal-2019-findings}, RWTH-Phoenix Weather 2014~\cite{koller15:cslr}, BBC-Oxford British Sign Language~\cite{Albanie2020bsl1k,Albanie2021bobsl}, and SMILE Swiss sign language dataset~\cite{ebling18} (respectively, from left to right). The full-body skeleton and hands mesh are on the right of each image. None of the demonstrated datasets was part of the model training. }
    \label{fig:qualitative}
\end{figure*}
\begin{table}
\centering
\setlength{\tabcolsep}{16pt}
\renewcommand{\arraystretch}{0.93}
\begin{tabular}{lc|c}\rowcolor{tablegray}
\hline
\multicolumn{1}{c}{} & \multicolumn{2}{c}{MPJPE (mm)}\\
\rowcolor{tablegray}
\multicolumn{1}{c}{\multirow{-2}{*}{Approach}} & \multicolumn{1}{c}{\textsc{RHD}} & \multicolumn{1}{c}{\textsc{STB}}\\ \hline
Zimmerman~{\etal}~\cite{zb2017hand} & 30.42 & 8.68 \\
Chen~{\etal}~\cite{Chen2018GeneratingRT} & 24.20 & 10.95 \\
Yang~{\etal}~\cite{disentagling_hands_linlin} & 19.95 & 8.66 \\
Spurr~{\etal}~\cite{spurr_cross-modal} & 19.73 & 8.56 \\
Moon~{\etal}~\cite{Moon_2020_ECCV_InterHand2.6M} & 20.89  & 7.95 \\
Gao~{\etal}~\cite{3d_interacting_gao} & 17.40  & 6.92 \\
Ours  & 16.98 & 7.56 \\ 
Ours$_{\text{D}}$  & {\bf 16.79} & 6.81 \\
Ours$_{\text{D}^*}$  & {\bf 16.79} & 6.47 \\\hdashline
Ours$_{\text{D}^*}$ (temporal) & - & {\bf 6.17} \\ \hline
\end{tabular}
\vspace{-0.2cm}
\caption{The table reports a comparison evaluation of~{\sota} methods and the proposed hand model estimators for \textsc{RHD}~\cite{zb2017hand} and \textsc{STB}~\cite{zhang_3d_hand_pose} datasets. The average per joint position errors (in mm) are reported in columns, and the lowest errors are highlighted in bold.
The (D) notation corresponds to the accuracy of a plain diffusion model, and (D$^*$) denotes the diffusion model with IK MLP.
After the dashed line is the evaluation of a temporal diffusion model. 
}
\label{table:rhd}
\end{table}


\begin{table}
\centering

\setlength{\tabcolsep}{4.pt}
\renewcommand{\arraystretch}{0.93}
\begin{tabular}{lrrr}\rowcolor{tablegray}
\hline
\multicolumn{1}{c}{} & \multicolumn{3}{c}{MPJPE (mm)}\\
\rowcolor{tablegray}
\multicolumn{1}{c}{\multirow{-2}{*}{Approach}} & \multicolumn{1}{c}{\textit{H}} & \multicolumn{1}{c}{\textit{M}} & \multicolumn{1}{c}{\textit{H}+\textit{M}} \\ \hline
Moon~{\etal}~\cite{Moon_2020_ECCV_InterHand2.6M}   &     10.42/13.05                   &   12.56/18.59                      &      12.16/16.02                    \\
Gao~{\etal}~\cite{3d_interacting_gao} & 9.10/12.82 & - & - \\ 
\multicolumn{2}{l}{Hampali~{\etal}~\cite{Hampali_2022_CVPR_Kypt_Trans}} - & - & 10.99/14.34 \\
\multicolumn{2}{l}{Fan~{\etal}~\cite{fan2021digit}} - & - & 11.32/15.57 \\
\multicolumn{2}{l}{Yu~{\etal}~\cite{yu2023acr}} - & - & {\bf 6.09}/{\bf 8.41} \\

Ours   &    9.09/12.61 &     13.37/19.06 &    11.98/16.04                      \\
Ours$_{\text{D}}$   &  {\bf 8.10}/{\bf 11.39}  & 11.97/18.58 & 10.44/14.81  \\ 
Ours$_{\text{D}^{*}}$   &  8.12/{\bf 11.39}  & {\bf 11.92}/{\bf 18.48}  & 10.43/14.78  \\ \hline

\end{tabular}

\vspace{-0.2cm}
\caption{The~\emph{state-of-the-art} comparison on \textsc{InterHand2.6M} dataset~\cite{Moon_2020_ECCV_InterHand2.6M}. The mean per joint position errors (in mm) for images of a single hand and interacting hands (separated by a slash symbol) are reported for human (\textit{H}),  machine (\textit{M}), and both (\textit{H}+\textit{M}) test annotations where the models were trained on the corresponding training sets. The Ours, Ours$_{\text{D}}$, and Ours$_{\text{D}^{*}}$ mark the baseline, diffusion, and diffusion + IK MLP models respectively.}
\label{table:interhand}
\end{table}

\vspace{-0.15cm}
The hand pose models for 3D estimation from a single RGB image were evaluated on three publicly available benchmark datasets.
First, the Rendered Hand pose Dataset (\textsc{RHD})~\cite{zb2017hand} contains synthetically generated images of humans with 3D hand skeletons provided.
It has 41258 training and 2728 testing samples with 20 different characters performing 39 actions.
The comparison to the {\sota} methods is shown in table~\ref{table:rhd}.
The proposed baseline and diffusion model have the lowest error, whereas the diffusion model with an additional IK module shows the best performance.
While the error-wise improvement of the IK model may not be substantial, the visual realism of the hand representation is notably enhanced (see Figure~\ref{fig:denoising_steps}).

The second, Stereo Hand pose Benchmark dataset (\textsc{STB})~\cite{zhang_3d_hand_pose} includes 18000 stereo reconstructed 3D hand poses, where 15000 images are used for training and 3000 for testing.
Results are reported in table~\ref{table:rhd} where the proposed models have the best accuracy, similarly, the diffusion model outperforms the baseline model.

The third, \textsc{InterHand2.6M}~\cite{Moon_2020_ECCV_InterHand2.6M} contains 2.6 million images of single and interacting hands including 3D poses triangulated from multiple views.
The 2D hand detections for \textsc{InterHand2.6M} were obtained by either manual annotation (H) or an automatic annotation tool (M), therefore the dataset can be divided into two parts depending on annotation type.
The results for different partitions of this dataset are shown in table~\ref{table:interhand}.
While the proposed model outperforms the baseline and recent methods, it should be noted that Yu~{\etal}~\cite{yu2023acr} have explicitly designed a model to address hands' interactions, whereas our method primarily focuses on a single hand, which shows superior accuracy on single hand datasets such as~\textsc{RHD} and~\textsc{STB}.

We used the evaluation script from~\cite{spurr_cross-modal} to compute model accuracy for \textsc{STB} and \textsc{RHD} datasets, and the script from~\cite{Moon_2020_ECCV_InterHand2.6M} for evaluation of \textsc{InterHand2.6M} dataset.
The results of other approaches were taken from the original papers.


\subsection{Temporal smoothing}
\begin{figure}
    \centering
    \includegraphics[width=0.19\linewidth]{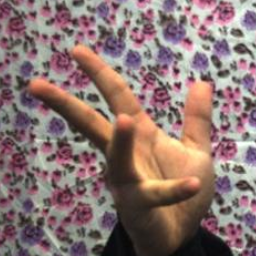}
    \includegraphics[width=0.19\linewidth]{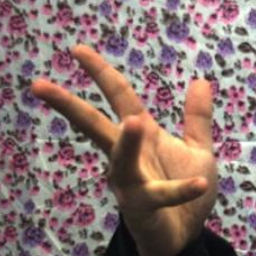}
    \includegraphics[width=0.19\linewidth]{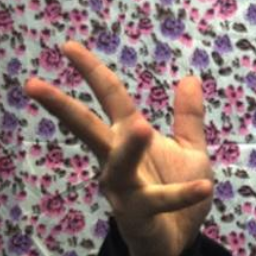}
    \includegraphics[width=0.19\linewidth]{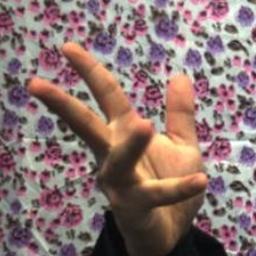}
    \includegraphics[width=0.19\linewidth]{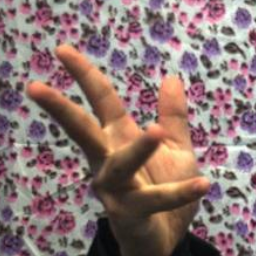}
    \includegraphics[width=0.99\linewidth]{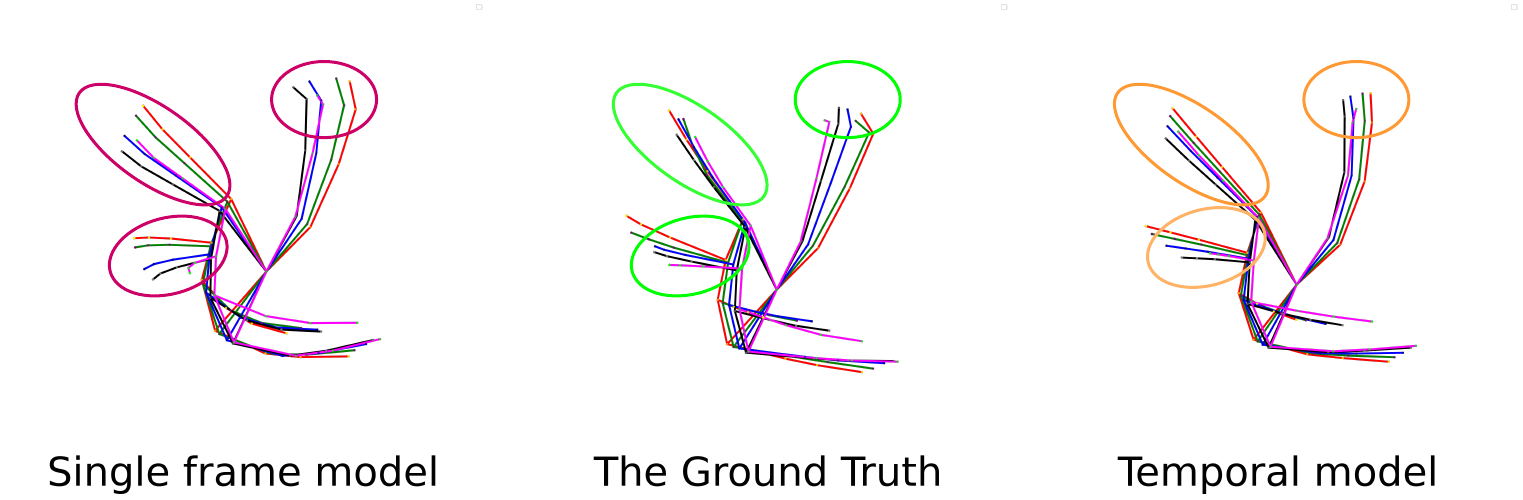}
    \vspace{-0.30cm}
    \caption{The top five images show a hand motion over consecutive frames. The bottom left figure shows 3D poses (centralized to the wrist joint) independently predicted by a single frame model, the middle figure shows the ground poses used for supervision, and the right figure shows the output of the temporal model. The circles highlight significant changes in the estimation. The input images were taken from the \textsc{STB} dataset~\cite{zhang_3d_hand_pose}.}
    \label{fig:temp_model}
\end{figure}
\vspace{-0.15cm}
The temporal transformer model was trained and quantitatively evaluated on the validation set of the \textsc{STB} dataset, which is the only dataset that has consecutive image frames available.
Using the temporal window of five frames, the model achieved a 4.6\% performance increase versus the accuracy of the diffusion model in table~\ref{table:rhd}.
The qualitative evaluation is demonstrated in Figure~\ref{fig:temp_model}, where the temporal model shows a smoother estimate with less jittering compared to a single-frame model.

\subsection{Qualitative evaluation}
\vspace{-0.2cm}
We evaluated the baseline model trained on a new partition of the SMILE dataset~\cite{ebling-etal-2018-smile}, which contains several million hand images.
For a qualitative evaluation, we randomly selected images from different sign language datasets that were not part of the training.
The MediaPipe~\cite{mediapipe} 2D detector was employed to get image points of a human to uplift the 3D body pose and localize hands in images.
The full-body pose estimation with a hand mesh is shown in Figure~\ref{fig:qualitative}.
It can be seen that the model has good generality across various images and could be used for sign-language tasks.


\section{Conclusions}
\vspace{-0.2cm}
This paper presents a novel 3D hand pose estimator from a single image.
The proposed method uses the denoising diffusion model to learn a hand distribution from images conditioned on ResNet features. This allows us to predict 3D structure from image features. 
Additionally, it exploits the skeletal structure of the hand to parameterize and constrain the 3D pose and enforce realistic estimation.
The baseline method reaches {\sota} results on multiple benchmark datasets, while the diffusion model further improves the accuracy.
To enforce temporal smoothness and remove jittering across frames, we introduce a temporal Transformer model which is applied to a consecutive sequence of frames providing further gains.
The approach was qualitatively evaluated on different sign language datasets not used in the training and demonstrates excellent generality.

\section{Acknowledgement}
\vspace{-0.2cm}
This work was supported by the EPSRC project ExTOL (EP/R03298X/1), SNSF project ’SMILE II’ (CRSII5 193686), European Union’s Horizon2020 programme (’EASIER’ grant agreement 101016982) and the Innosuisse IICT Flagship (PFFS-21-47). This work reflects only the authors view and the Commission is not responsible for any use that may be made of the information it contains.

{\small
\bibliographystyle{ieee_fullname}
\bibliography{egbib}
}

\end{document}